# Synthetic Image Augmentation for Damage Region Segmentation using Conditional GAN with Structure Edge

Takato Yasuno[*1] Michihiro Nakajima[*1], Seiji Sekiguchi[*1],

Kazuhiro Noda[*1], Kiyoshi Aoyanagi[*1], Sakura Kato[*1]

[*1] Yachiyo Engineering, Co., Ltd.

Recently, social infrastructure is aging, and its predictive maintenance has become important issue. To monitor the state of infrastructures, bridge inspection is performed by human eye or bay drone. For diagnosis, primary damage region are recognized for repair targets. But, the degradation at worse level has rarely occurred, and the damage regions of interest are often narrow, so their ratio per image is extremely small pixel count, as experienced 0.6 to 1.5 percent. The both scarcity and imbalance property on the damage region of interest influences limited performance to detect damage. If additional dataset of damaged images can be generated, it may enable to improve accuracy in damage region segmentation algorithm. We propose a synthetic augmentation procedure to generate damaged images using the image-to-image translation mapping from the tri-categorical label that consists the both semantic label and structure edge to the real damage image. We use the Sobel gradient operator to enhance structure edge. Actually, in case of bridge inspection, we apply the RC concrete structure with the number of 208 eye-inspection photos that rebar exposure have occurred, which are prepared 840 block images with size 224 by 224. We applied popular per-pixel segmentation algorithms such as the FCN-8s, SegNet, and DeepLabv3+Xception-v2. We demonstrates that re-training a dataset added with synthetic augmentation procedure make higher accuracy based on indices the mean IoU, damage region of interest IoU, precision, recall, BF score when we predict test images (236 words).

## 1. Introduction

### 1.1 Related GAN Studies for Accuracy

Recently, social infrastructure is aging, and its predictive maintenance has become an important issue. In order to monitor the state of infrastructures, the inspection is performed manually by human eye or automatically by drone. And locations of damage for screening are detected, in addition primary damage region for repair targets are segmented per-pixel. After these task, we often select critical repair targets for predictive maintenance. Here, we are required to accurately inspect them. However, deterioration has rarely occurred, and the damage regions are often narrow, so their ratio within one image is extremely small, i.e., we experienced cases such as 0.6 and 1.5 percent. Such an imbalance property of the class weight of ROI-damage toward background influences constrained performance to improve accuracy. If this sparse damaged region can be duplicated and an additional dataset of images and labels can be generated, it will be possible to achieve stable training process and improved accuracy in damage region segmentation task.

In the field of social infrastructure inspection, there are related works to detect their damages such as object detection task [Gopalakrishnan 2018] and semantic segmentation researches [Guillanmon 2018]. The damaged class is rare event and the dataset including that is always imbalance, so the number of rare class images is very small. The more damaged, the less event occurred to collect images. Because of this scarcity of damaged data, it is difficult to improve the accuracy of damaged interest in social infrastructure monitoring and inspection. Especially, such a damaged interest images deteriorated is scarce event, the useful dataset that consists damage region of interest took much time consuming. This is one of hurdle to overcome our underlying problems for social infrastructure aging detection for data mining from supervised learning approaches. Instead, we proposes a synthetic augmentation procedure in order to generate inspection images with damage interest from unsupervised approach.

Since 2014, the original generative adversarial network (GAN) paper is cited more than 9,000 times to date (July 2019). Starting from GAN's invention in 2014, the field of GAN has been growing exponentially over 360 papers [Hindupur 2017]. GANs may be used for many applications, not just fighting breast cancer or generating human faces, but also 62 other medical GAN applications published through the end of July 2018 [Kazeminia 2018]. Using DAGAN for data augmentation, they achieved a significant improvement in classification accuracy compared to the baseline of standard data augmentation only [Frid-Adar 2018]. They added synthetic data produced by their DCGAN, then the classification performance improved from around 80% to 85%, demonstrating the usefulness of GANs.

In order to overcome the scarcity of rare class imageless including damage region of interest, we expect the usefulness of synthetic augmentation added the rare class images. We call "Synthetic Augmentation". However, in the field of semantic segmentation task for monitor social infrastructure, it is not clear that synthetic augmentation can improve the segmentation accuracy. We demonstrate several training and test added synthetic augmentation using L1-Conditional GAN.

### 1.2 Synthetic Image Augmentation using cGAN

We think that approaches for generating a damage image include 1) reproducing the already acquired damage image (Similar augmentation), and 2) generating a future image degraded from the current damage grade (What-if degradation)

Contact: Takato Yasuno, RIIPS on 5-20-8, Asakusabashi, Taito-ku, Tokyo, 111-8648, tk-yasuno@yachiyo-eng.co.jp.

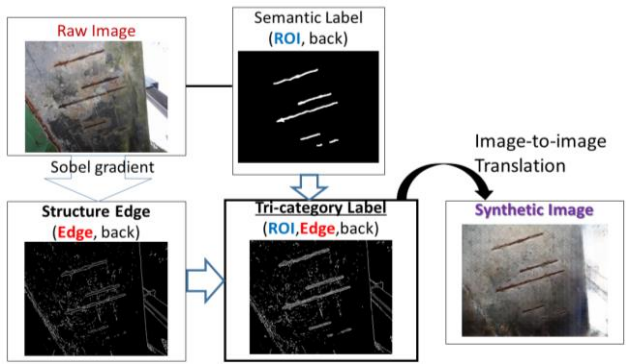

Figure 1: Synthetic augmentation method using image-to-image translation mapping semantic label with structure edge into image.

and 3) what-if newer damage that does not yet exist (what-if newer). Here, 1) is close to data augmentation that has been performed as standard in supervised learning by rotation, X/Y translation, scaling, and so forth. This is useful for giving variants to the features of the acquired image, increasing variations, accelerating learning, and increasing generality.

In case of 2), it is possible to simulate the situation where the deterioration has progressed several years ahead of the current state. Degraded state that has not yet been experienced, but generates an image of the state that has progressed one rank deterioration or the worst image when the management level is low, exceeds the scope where the supervised data exists. This is an attempt to eliminate any blind spots in the supervised learning. 3) is an approach that was not possible with supervised learning based on the experience data. Even in case of social infrastructures that has not deteriorated, a new damaged image can be generated in order to prepare for future deterioration, and even if it has not yet been experienced, it enables to imagine a degraded future image.

However, it is necessary to have a reality about where and how much damage occurs in the infrastructure. This means that after acquiring images of potential damage throughout the life cycle of social infrastructure. It is necessary to design a new possible damage scenario and place the damage at the possible position. It is necessary to generate a new damaged image with ethics in order to make the manager uneasy about the damaged image without reality. This paper try to expand the damaged image using the most basic method 1) of reproducing the current damaged image.

## 2. Generative Damage Augmentation

### 2.1 Damage Segmentation Architectures

In order to recognize the damage region of interest for social infrastructure, semantic segmentation algorithms are useful. We propose a synthetic augmentation method to generate fake images and labels using the L1-conditional GAN (pix2pix) to translate a label image with structure edge to a damaged image. We apply several existing per-pixel segmentation task based on transfer learning such as the Fully Convolutional Network (FCN) [Long 2015] based on AlexNet and VGG16 with different skip connections that we call the type 8s, 16s, and 32s, furthermore the SegNet [Badrinarayanan 2016], and the dense convolution network such as the DeepLabv3+ResNet18, ResNet50, Xception-v2 [Chen 2018]. We compare the trained segmentation accuracy using initial dataset with the re-trained segmentation accuracy using synthetic augmentation added generated images. We evaluate the both task performance to compute the similarity indexes between the ground truth damage region of interest (ROI) and the predicted region. Exactly, we compute the mean Intersection of Union (mIoU), class-IoU that consists the ROI and background. In order to analyze the property of synthetic augmentation, we compute the precision, recall and BF score. Therefore, using these existing segmentation architectures, we get some knowledge whether our method of synthetic augmentation can improve their segmentation accuracy or not.

### 2.2 Synthetic Augmentation from Semantic Label with Structure Edge to Image

To train the DCGAN, we need more than 500 images and also they should have their stable angle. In the infrastructure deterioration process, a progressed damage is rare event and it is not easy to collect their damaged images more than even several hundreds. The eye-inspection view has various angle according to each field to monitor their social infrastructures. On the other hand, the image-to-image translation is possible for training a paired image dataset even with various inspection angle. This paper propose a synthetic augmentation method using L1-Conditional GAN (pix2pix). The original pix2pix paper translated form the input of edge images to shoe images [Isola 2018]. And using the CamVid dataset, they translated from the semantic label to photo. However in case of damage images, we could not success such a naive translation. As shown Figure 1, this paper proposes the semantic label with structure edge as input of tri-categorical labels. This augmented label consists damage-ROI, enhanced structure edge, and background. We tried several edge detection method such as Gradient operators (Roberts, Prewitt, Sobel), Laplacian of Gaussian (LoG), Zero crossing, Canny edge and so forth [Gonzalez 2018]. We selected the Sobel gradient operator that is a method of finite differences between the pixel's function value and that of its right (or left) neighbor gives gradient at that pixel. This operator is robust to noise. This paper propose the Sobel detection in order to extract the background edge from eye-inspection photo. It is possible the structure feature of concrete parts that consists social infrastructure such as bridge. We combine the both semantic label and structure edge produced by the Sobel edge detection into three class categorical label. We train the mapping from the semantic label with structure edge to damaged image.

Thus, we summarize a synthetic augmentation step as follows. First, we train one of semantic segmentation task using the initial dataset including with eye-inspection images and semantic label. Second, we apply a synthetic augmentation method mapping to generate fake images using L1-Conditional GAN from combined semantic label with structure edge. Third, we re-train another semantic segmentation task using the both initial dataset and synthetic augmented dataset added their generated inspection images. The number of dataset is two times compared with the initial dataset, so as to extend an opportunity to learn the damage feature between real inspection photos and synthetic images.

## 3. Applied Results

### 3.1 Bridge inspection dataset

This paper focuses on the one of social infrastructure inspection, exactly concrete bridge eye-inspection dataset. Actually, bridge eye-inspection photos has lower resolution and also heterogeneous size range from 360 pixels to 1,500. We select the ROI-rich images from around 20 thousands eye-inspection photos. We got the part of RC concrete structure with the number of 208 inspection photos that rebar exposure have occurred. This extracted rate is only one percent, so the rebar exposure is also rare event. We annotated ROI and background class labels over each raw image for semantic segmentation task. Without loss of resolution, in order to keep the pixel feature data we prepare to extract 998 block images unified with size 224 by 224. Furthermore, we delete small size block images less than 128 pixel, and unusable images without damage-ROI. After these cleansing, we have a dataset with number of 840 images. We compute the class weight that the damage-ROI weight is 16.07 and background weight is 0.51 divided by median pixel count.

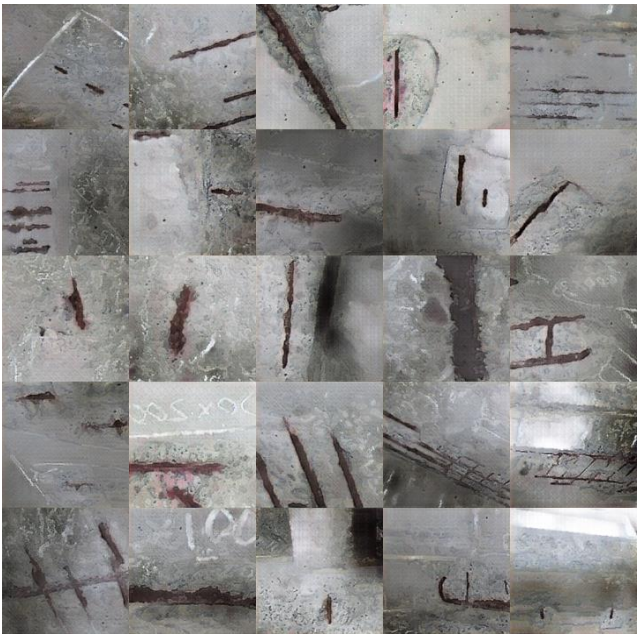

Figure 2: Generated images using synthetic augmentation using L1-Conditional GAN in case of rebar exposure at concrete bridge.

### 3.2 Synthetic Augmentation and Generated Images

We applied the L1-Conditional GAN that carried out image-to-mage translation from tri-categorical labels combined between the semantic label and structure edge by Sobel detection into the real training dataset 840 block images. We trained 200 epoch that took 13 hours. The L1 penalty coefficient is 100 at loss function. Figure 2 shows the generated images by synthetic augmentation.

### 3.3 Accuracy Comparison Without-With Deep Fake

We trained initial dataset using RMSProp optimizer 3epoch with mini batch 16 to 32, around 10 to 20 thousands iterations. We did standard augmentation random crop extraction multiplied 64 crops with unit size 224 by 224. We partitions the dataset whose weight train versus test is 95 : 5, each number of dataset consists 798 and 42. Next, we trained synthetic augmentation with the number of 1680, so as to compare the initial result where we set the same 3 epoch, partition weight 95 : 5, and mini batch 16 to 32, where the FCN-8s, SegNet needs much memory.

Tabel 1 shows the trained results consist each running time, mean IoU, class-IoU (rebar exposure, background). The running time took around two times more than the initial dataset, because we added the generated images using synthetic augmentation over the initial dataset. The FCN-AlexNet using synthetic augmentation outperform the initial trained accuracy to evaluate value of the mean IoU and rebar exposure IoU. And also, FCN-8s, FCN-16s, and SegNet-VGG16 have improved mean IoU and rebar exposure-IoU more than the initial trained one. Furthermore, two dense convolutional networks indicated high performance, these are the DeepLabv3+ResNet50 and Xception. Therefore, we demonstrated that synthetic augmentation using L1-Conditional GAN enable to improve the segmentation accuracy, though it is not always the better off.

Table 1: Trained and test predicted results of intersection of union.

| architecture | dataset | runing time | mean IoU | ROI-IoU | background-IoU |
|---|---|---|---|---|---|
| **FCN-AlexNet** | initial | 49m | 0.5376 | 0.1346 | 0.9405 |
| | synthetic augment. | 100m | **0.5874** | **0.2162** | 0.9585 |
| **FCN-8s** | initial | 210m | 0.6289 | 0.2778 | 0.9801 |
| | synthetic augment. | 336m | **0.7367** | **0.4851** | 0.9883 |
| **FCN-16s** | initial | 190m | 0.6175 | 0.2612 | 0.9738 |
| | synthetic augment. | 332m | **0.6759** | **0.3720** | 0.9797 |
| FCN-32s | initial | 167m | 0.5796 | 0.1963 | 0.9629 |
| | synthetic augment. | 336m | 0.5723 | 0.1999 | 0.9446 |
| **SegNet-VGG16** | initial | 274m | 0.6574 | 0.3263 | 0.9884 |
| | synthetic augment. | 480m | **0.8135** | **0.6344** | 0.9926 |
| DeepLabv3 +ResNet18 | initial | 66m | 0.6951 | 0.4044 | 0.9857 |
| | synthetic augment. | 182m | 0.7137 | 0.4447 | 0.9826 |
| **DeepLabv3 +ResNet50** | initial | 170m | 0.7289 | 0.4686 | 0.9892 |
| | synthetic augment. | 324m | **0.8005** | **0.6082** | 0.9928 |
| **DeepLabv3 +Xception** | initial | 275m | 0.6531 | 0.3255 | 0.9807 |
| | synthetic augment. | 549m | **0.7902** | **0.5886** | 0.9917 |

### 3.4 Predict Test Images beyond Initial Dataset

In order to evaluate more general accuracy, we searched and downloaded another rebar exposure images including concrete infrastructure such as bridge and building. We tried to predict these newer test images with the number of 40, where we prepare center crop procedure located on some rebar exposure. Table 2 shows the predicted results applied on the initial trained networks and another trained network using synthetic augmentation. Especially, our synthetic augmentation procedure can perform higher precision from the viewpoint of the both rebar exposure and background. Figure 3 shows the overlay of two labels between the ground truth region of damage interest and the predicted region produced by the segmentation task. The top images stands for the initial dataset based prediction, in contrast the bottom images denotes the synthetic segmented prediction. Using our synthetic augmentation procedure, the target region of interest are approaching to the ground truth of rebar exposure. Exactly, we demonstrated five improved architectures such as the FCN-AlexNet, FCN-8s, FCN-16s, SegNet-VGG16, and DeepLabv3+Xception. On the other hand, the recall sometimes made a little bit better off. In result, our synthetic augmentation procedure can improve the precision accuracy at the semantic segmentation task.

Table 2: Test prediction beyond initial dataset, precision, recall and BF.

| architecture | dataset | precision | | recall | | BF score | |
|---|---|---|---|---|---|---|---|
| | | ROI | background | ROI | background | ROI | background |
| **FCN-AlexNet** | initial | 0.1252 | 0.5937 | 0.1546 | 0.5494 | 0.1296 | 0.5573 |
| | synthetic augment. | **0.1861** | **0.6975** | 0.1506 | **0.5721** | **0.1532** | **0.6176** |
| **FCN-8s** | initial | 0.2892 | 0.6347 | 0.3787 | 0.7085 | 0.2951 | 0.6575 |
| | synthetic augment. | **0.3757** | **0.7182** | **0.3870** | 0.7089 | **0.3600** | **0.7031** |
| **FCN-16s** | initial | 0.1742 | 0.6374 | 0.1844 | 0.6179 | 0.1688 | 0.6207 |
| | synthetic augment. | **0.2426** | **0.7031** | **0.2163** | **0.6238** | **0.2134** | **0.6524** |
| FCN-32s | initial | 0.1622 | 0.7099 | 0.1497 | 0.5527 | 0.1465 | 0.6094 |
| | synthetic augment. | 0.1298 | 0.6509 | 0.1254 | 0.5298 | 0.1187 | 0.5738 |
| **SegNet-VGG16** | initial | 0.2979 | 0.6560 | 0.4181 | 0.7202 | 0.3267 | 0.6798 |
| | synthetic augment. | **0.3912** | **0.7063** | **0.4854** | **0.7457** | **0.4124** | **0.7186** |
| DeepLabv3+ResNet18 | initial | 0.3391 | 0.7562 | 0.2804 | 0.7021 | 0.2916 | 0.7203 |
| | synthetic augment. | 0.1958 | 0.5798 | 0.2580 | 0.6361 | 0.2090 | 0.5973 |
| DeepLabv3+ResNet50 | initial | 0.3981 | 0.7598 | 0.3352 | 0.7018 | 0.3451 | 0.7227 |
| | synthetic augment. | 0.3937 | **0.7918** | 0.2587 | 0.6768 | 0.2849 | 0.7189 |
| **DeepLabv3+Xception** | initial | 0.2906 | 0.6491 | 0.2988 | 0.6807 | 0.2714 | 0.6509 |
| | synthetic augment. | **0.3982** | **0.7327** | **0.3197** | 0.6701 | **0.3130** | **0.6839** |

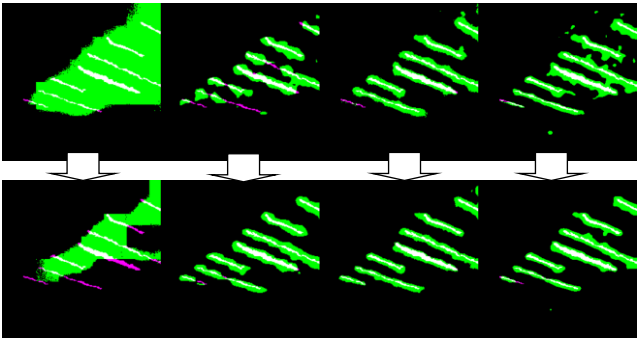

Figure 3: Overlay between ground truth and predicted mask, to compare the initial segmentation (top) with the synthetic augmentation (bottom). From left to right, we show our prediction results as follows : FCN-AlexNet, FCN-8s, SegNet-VGG16, DeepLabv3+Xception.

## 4. Concluding Remarks

This paper proposes a synthetic augmentation procedure using L1-Conditional GAN. This is an image-to-image translation algorithm which is mapping from tri-categorized labels that consists ROI, structure edge, and background, into real damaged image. We propose a Sobel edge to extract the feature of structure edge from eye-inspection photo. Therefore, we demonstrated that our synthetic augmentation procedure using L1-Conditional GAN, which enable to improve the segmentation accuracy, though it is not always the better off. Especially, our synthetic augmentation procedure can perform higher precision from the viewpoint of the both rebar exposure and background. Using our synthetic augmentation procedure, the target region of interest were approaching to the ground truth of rebar exposure. Exactly, we demonstrated architectures such as the FCN-AlexNet, FCN-8s, SegNet-VGG16, and DeepLabv3+Xception and so forth.

Furthermore, we will challenge to develop a pioneer architecture for social infrastructure health monitoring and asset management. This paper focused on the road bridge, in future we would like to increase opportunities to apply dam and river. In future, another synthetic augmentation due to newer occurrence of not yet experienced damage and what-if degradation will be series issue using Cycle/StyleGAN. For more general purpose application, while maintaining the reality and ethically paying attention to practical concerns for infrastructure managers.

[Acknowledgments] We would thank Mr. S. Kuramoto and Mr. T. Fukumoto for providing us information for GAN frameworks.

### References

[Gopalakrishnan 2018] Gopalakrishnan, K., Gholami, H. et al. : Crack Damage Detection in Unmanned Aerial Vehicle Images of Civil Infrastructure using Pre-trained Deep Learning Model, International Journal for Traffic and Transport Engineering, 8(1), pp.1-14, 2018.

[Ricard 2018] Ricard, W., Silva, L. et al. : Concleto Cracks Detection based on Deep Learning Image Classification, MDPI Proceedings, 2, 489, pp.1-6, 2018.

[Guillanmon 2018] Guillamon, J.R. : Bridge Structural Damage Segmentation using Fully Convolutional Networks, Universitat Politecnica de Catalunya, 2018.

[Yasuno 2019] Yasuno, T. : Sparse Damage Per-pixel Prognosis Indices via Semantic Segmentation, 33th Journal of Society for Artificial Intelligence, 3B3-E-2-05, 2019.

[Hindupur 2017] Hindupur, A. : The GAN Zoo, https://github.com/hindupuravinash/the-gan-zoo.

[Kazeminia 2018] Kazeminia, S. et al. : GANs for Medical Image Analysis, https://arxiv.org/pdf/1809.06222.pdf.

[Frid-Adar 2018] Frid-Adar, M., Diamant, I. et al : GAN-based Synthetic Medical Image Augmentation for increased CNN Performance in Lesion Classification, CVPR, 2018.

[Long 2015] Long, J. et al: Fully Convolutional Networks for Semantic Segmentation, CVPR, pp3431-3440, 2015.

[Badrinarayanan 2016] Badrinarayanan, V., Kendall, A. et al., SegNet: Deep Convolutional Encoder-Decoder Architecture for Image Segmentation, ArXiv:1511.00561v3, 2016.

[Chen 2018] Chen, L-C., Zhu, Y., Papandreou, G. et al : Encoder-Decoder with Atrous Separable Convolution for Semantic Image Segmentation, arXiv:1802.02611v3.

[Isola 2018] Isola, P. et al. : Image-to-image Translation with Conditional Adversarial Network, CVPR, 2017.

[Gonzalez 2018] Gonzalez, R.C., Woods, R.E. : Digital Image Processing, 4th Global Edition, Pearson. (2020.March 4)